\documentclass[runningheads]{llncs}

\usepackage[T1]{fontenc}
\usepackage{caption}

\usepackage{graphicx}
\usepackage{hyperref}
\usepackage{color}

\usepackage{algorithm,caption}
\usepackage{algpseudocode}

\usepackage{adjustbox}
\usepackage{siunitx,booktabs}
\usepackage{xcolor,colortbl}
\usepackage{diagbox}
\usepackage{multirow}
\usepackage{amsmath}
\usepackage{amssymb}
\usepackage{hyperref}
\usepackage{bbding}

\usepackage{graphicx}

\DeclareRobustCommand\onedot{\futurelet\@let@token\@onedot}
\def\@onedot{\ifx\@let@token.\else.\null\fi\xspace}

\def\ie{\emph{i.e}\onedot}

\def\etal{\emph{et al}\onedot}

\newcommand{\loss}[1]{\mathcal{L}_\text{#1}}

\newcommand{\repeatthanks}{\textsuperscript{\thefootnote}}

\begin{document}
\title{Adaptive Personlization in Federated Learning for Highly Non-i.i.d. Data}

\author{Yousef Yeganeh\thanks{Equal Contribution}\inst{1} \and Azade Farshad\repeatthanks \inst{1} \and
Johann Boschmann\inst{1} \and Richard Gaus \inst{1} \and Maximilian Frantzen\inst{1} \and
Nassir Navab\inst{1,2}
}

\authorrunning{Y. Yeganeh et al.}
\institute{Technical University of Munich, Munich, Germany \and
Johns Hopkins University, Baltimore, USA
}

\authorrunning{Y. Yeganeh et al.}

\maketitle              
\begin{abstract}
Federated learning (FL) is a distributed learning method that offers medical institutes the prospect of collaboration in a global model while preserving the privacy of their patients. Although most medical centers conduct similar medical imaging tasks, their differences, such as specializations, number of patients, and devices, lead to distinctive data distributions. Data heterogeneity poses a challenge for FL and the personalization of the local models. In this work, we investigate an adaptive hierarchical clustering method for FL to produce intermediate semi-global models, so clients with similar data distribution have the chance of forming a more specialized model. Our method forms several clusters consisting of clients with the most similar data distributions; then, each cluster continues to train separately. Inside the cluster, we use meta-learning to improve the personalization of the participants' models. We compare the clustering approach with classical FedAvg and centralized training by evaluating our proposed methods on the HAM10k dataset for skin lesion classification with extreme heterogeneous data distribution. Our experiments demonstrate significant performance gain in heterogeneous distribution compared to standard FL methods in classification accuracy. Moreover, we show that the models converge faster if applied in clusters and outperform centralized training while using only a small subset of data.
\end{abstract}
\section{Introduction}\label{intro}
\begin{figure}[htb]
    \centering
    \includegraphics[width=\linewidth]{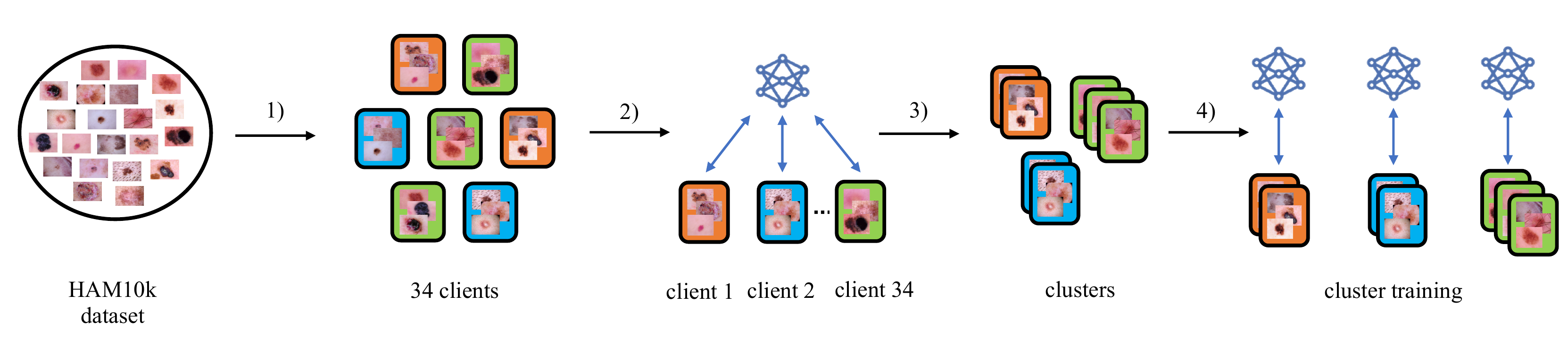}
    \caption{\textbf{Method Overview}. Our pipeline has four steps: 
    1) splitting the dataset, 
    2) training all clients with FedAvg for predefined rounds,
    3) clustering the clients based on the latest model update,
    4) performing FedAvg or FedAP on each separate cluster. 
    }
    \label{fig:pipeline}
\end{figure}

Deep learning models outperform classic techniques for pathological diagnoses in medical imaging tasks \cite{madabhushi2016image,farshad2022upsilon,aliakbari2015simulation}; however, their performance highly depends on the training data. Unavailability of medical imaging data due to privacy concerns, along with data heterogeneity, can negatively impact the representativity of the model. Federated learning (FL) tackles both of these challenges\cite{li2020federated}. A federated setting consists of multiple clients and a server; local clients send their models to the server, and the server aggregates them and produces a global model\cite{mcmahan2017}. Weighted aggregation in FL aims to improve performance in favor of a global model; in turn, parts of the data distribution that potentially have distinctive features are considered outliers, e.g., since hospitals often outweigh specialized centers in terms of the number of patients, the global model tends to represent hospitals' distribution, so the prospect of collaboration among institutes with similar data distribution is neglected. To improve personalization, we pose FL as a meta learning problem \cite{finn2017,nichol2018}. We demonstrate that clustering and applying a meta learning scheme improve personalization, preserve more specialized data, and enhance the convergence of the model.
FedAvg (federated averaging) \cite{mcmahan2017} was proposed as one of the first FL algorithms and has been used as a standard benchmark. Meta learning, or learning to learn \cite{vanschoren2018meta}, is learning how to efficiently solve new tasks from a set of known tasks \cite{finn2017}. FL clients can be interpreted as meta learning tasks since each client's data distribution equals a different problem; hence, meta learning ideas have been successfully applied to FL \cite{jiang2019improving,fallah2020personalized,chen2019federated,khodak2019adaptive}. Inspired by MAML\cite{finn2017}, clients with similar distributions are grouped in clusters as a set of tasks, so each cluster is redefined a separate FL problem. This improves the homogeneity of the data between the clients in their corresponding clusters. Although in FL data is not explicitly accessible, we used the value differences of model parameters between the clients from the latest round and the global model as a similarity measure between those clients. Briggs \etal explores a hierarchical clustering technique to group similar models  \cite{briggs2020federated}, and we utilized the same method for personalization of clients' models and combined it with our proposed Adaptive Personalization (FedAP) for the final training inside clusters. We kept the simplicity of FedAvg \cite{jiang2019improving}, but with two differences: 1) FedAP treats the aggregated global update as a meta-level gradient that can be used with a different optimizer to update the global model. Particularly, it introduces the new hyperparameter of a meta learning rate. 2) At the end of the training, FedAP personalizes the global model to each individual client. This can have a strong advantage in the context of FL \cite{fallah2020personalized,jiang2019improving}. An overview of our method is depicted in ~\autoref{fig:pipeline}. The main contributions of this work are as follows: 1) We propose FedAP, a new hierarchical clustering approach to perform adaptive personalization inside the clusters and each client. 2) Significant performance gain in terms of classification accuracy and decreasing the accuracy variance between different clients. 3) The code of this work will be publicly released upon its acceptance.

\section{Related work}
A study on live data of millions of users shows that significant improvements can be achieved by personalizing the learning rate and batch size to clients \cite{wang_federated_2019}. The FedOpt \cite{reddi_adaptive_2020} extends the FedAvg algorithm and implements personalization by introducing adjustable gradient update strategies for each client and server. Employing weight decay over training rounds on the server-side is shown to be required to lower the error on non-i.i.d. data \cite{li_convergence_2020}. \cite{arivazhagan_federated_2019} proposes to learn base layers globally while keeping classification layers private on the client-side. FedMD \cite{li_fedmd_2019} introduces a framework for individually designed by each client. The MOCHA \cite{smith_federated_2018} enables efficient meta-learning in the federated environment. Communication efficiency in federated learning is also addressed in \cite{sattler_robust_2019} suggesting a compression protocol based on quantization. Non i.i.d. data is shown to impact both the convergence speed and the final performance of the FedAvg algorithm \cite{li_convergence_2020,sattler_robust_2019}. \cite{zhao2018federated,li_convergence_2020} tackle data heterogeneity by sharing a limited common dataset. IDA \cite{yeganeh_inverse_2020} proposes to stabilize and improve the learning process by weighting the clients' updates based on their distance from the global model. Motivated from classical machine learning techniques \cite{li2020federated} introduces a weight regularization term to the local objective function to prevent the divergence between local and global models. Yue \etal \cite{yue_deep_2020} surveys on the data availability and heterogeneity in the medical domain, concluding that privacy restrictions and missing data pipelines block the full potential of Deep Learning which requires big data sets. The described advances in the field of federated learning can help overcome the challenge of medical data privacy and disseminate machine learning techniques in healthcare \cite{sheller2020federated}. Notably, data heterogeneity remains a significant hurdle to this development, so robust techniques towards non-i.i.d. data carry considerable future potential \cite{rieke2021future}.
\section{Method}\label{sec:method}
In this section, we present our approach to tackling the problem of non-i.i.d. data distribution in federated learning. First, we describe the optimization process in a federated setting and its development in adaptive personalization. Then, we define the original federated averaging scenario. Later, we explain the entire pipeline of our proposed federated adaptive personalization method, followed by how hierarchical clustering can be embedded in our approach.

\subsection{Definitions}
The global data distribution is denoted by $\mathcal{D}$, while $\mathcal{S} \sim \mathcal{D}$ is the sampled data points and $s_1, ... s_N = \mathcal{S}$. The data points are distributed across $M$ clients. Each client $i \in \{1, \dots, M\}$ only sees its local dataset, which is a subset $\mathcal{S}_i \sim \mathcal{D}_i$ of the global data and $s_1, ..., s_{N_i} = \mathcal{S}_i$. $\mathcal{D}_i$ being the client's local data distribution. With the global model parameters $\theta_{global}$, the overall optimization task can be defined as:
\begin{equation}
\label{eqn:globalLoss}
\min_{\theta_{global}} \loss{}(\mathcal{S}) = \min_{\theta_{global}} \frac{1}{M} \sum_{i=1}^{M} l_{\theta_{global}}(\mathcal{S}_i)
\end{equation}
where $\loss{}$ is the global loss function and $l$ the local loss function of each client.
We hypothesize that, while each distribution $\mathcal{D}_i$ is different, they can be clustered by similarity. Following Briggs \etal \cite{briggs2020federated}, we introduce a distinct model $\theta_{c}$ for each resulting cluster $c \in C$, where $C$ is the set of all clusters and $\theta_C$ is the set of all cluster model parameters.

We annotate the data distribution in each cluster by $\mathcal{D}_{c}$ with $\mathcal{S}_c \sim \mathcal{D}_c$ being the data points and $s_1,...,s_{N_c} = \mathcal{S}_c$. 
Integrating these clusters into \autoref{eqn:globalLoss}, we arrive at the following global and local loss function for each cluster $c$: 
\begin{align}
\label{eqn:clusterLoss}
\min_{\theta_{C}} \loss{}(\mathcal{S}) &= \min_{\theta_{C}} \frac{1}{|C|} \sum_{c \in C} l_{\theta_{c}}(\mathcal{S}_c) \\
l_{\theta_c}(\mathcal{S}_c) &= \frac{1}{|c|} \sum_{i \in c} l_{\theta_c}(\mathcal{S}_i)
\end{align}
Extending this model to a personalized version where each client has its own model results in \autoref{eqn:personalizedLoss}:
\begin{equation}
\label{eqn:personalizedLoss}
\min_{(\theta_1,..., \theta_n)} \loss{}(\mathcal{S}) = \min_{(\theta_1,..., \theta_n)} \frac{1}{N} \sum_{i=1}^{N} l_{\theta_{i}}(\mathcal{S}_i)
\end{equation}
\autoref{eqn:clusterLoss} and \autoref{eqn:personalizedLoss} both introduce additional degrees of freedom which allow to learn the sampled training data points from the respective distributions better than a single model as shown in \autoref{eqn:globalLoss}. We can sum up the results in the following form:
\begin{equation}
\label{eqn:lossInequality}
\min_{\theta_{global}} \loss{}(\mathcal{S}) \geq \min_{\theta_C} \loss{}(\mathcal{S})  \geq \min_{(\theta_1,..., \theta_M)} \loss{}(\mathcal{S})
\end{equation}

While \autoref{eqn:lossInequality} introduces the idea of each client learning its own model to reduce the loss on its distribution, the amount of data available for training is reduced. As theoretical works in deep learning show, the generalization error of models grows if the number of training samples shrinks. This can also be observed in practice and is reflected by the overfitting phenomenon.
To maintain a good generalization performance, we propose to train on all of the mentioned levels, gathering information about as many training samples as possible while reducing the problem complexity each model has to solve step by step. We start an initial training overall clients to find a global model $\theta_{global}$ which should learn basic features of the whole distribution. In the next step, we cluster the clients and begin with the training inside the clusters. The resulting cluster models $\theta_C$ are then personalized in the final step by performing local training on each client $i$, leading to the final personalized model $\theta_i$ for each client. \autoref{fig:pipeline} visualizes the whole pipeline.

\subsection{Federated Averaging}
In the following, we analyze the federated averaging (FedAvg) algorithm in more detail. If we have $M$ clients and $N_{i} = |\mathcal{D}_{i}|$ is the number of data samples of client $i$, the global and local loss functions take the following form 
\begin{equation}
f(\theta)=\sum _{i=1}^{M} \frac{N_i}{N}  F_i({\theta}) \quad \textrm{where} \quad F_i (\theta) = \frac{1}{N_i} \sum_{j} f_j (\theta)
\end{equation}
where, $F_{i}(\theta)$ is the local objective of each client and $f(\theta)$ is the global objective averaged over all clients.
Training with FedAvg consists of two main updating schemes.
Firstly, the clients are training locally with an initially distributed model on their local data for a predefined amount of epochs $e$. For each communication round a client updates its model parameter $\theta_i$ as follows,
\begin{equation}
\forall i, \  \theta_i^{t+1} \leftarrow \theta_i^t - \eta \cdot  \Delta \theta
\end{equation}
In a second step, the server computes a weighted average based on the number of data points of each client $i$ of all locally updated model parameters $\theta_{i}'s$: 
\begin{equation}
 \theta_{global}^{t+1} \leftarrow \sum_{i=1}^{M} \frac{N_i}{N} \theta_i^{t+1} 
\end{equation}

\subsection{Federated Adaptive Personalization} \label{meth:ml}
Our methodology (Federated Adaptive Personalization or FedAP) begins with the initialization of the global model parameters $\theta_{global}$. At the start of each of $k$ federated rounds, they are transferred to a batch of $n$ randomly selected clients, and local training is performed, yielding updated local model parameters $\theta_i$ in each client $i$. Next, $\theta$ is updated using an adaptive meta learning rate $\eta$ based on the current round number. The adaptive meta learning rate decreases linearly throughout the training.
\begin{equation}
    \theta_{global} \gets \theta_{global} + \eta \sum_{i = 1}^n \frac{N_i}{N} (\theta_i - \theta_{global})
\end{equation}
Following the last federated round, the local model parameters of all clients $i \in \{ 1, \dots, M \}$ are personalized by performing a fixed number of gradient optimization epochs on the local training data.

\subsection{Hierarchical Clustering}
The personalization of the local models can benefit more from clients that share more similarities in their data with each specific client. Therefore, we propose to perform hierarchical clustering after an initial phase of global federated learning, including all the clients. After performing FedAvg for a specified number of federated rounds, clients are clustered via hierarchical clustering according to their model updates in the current round. Each of the resultant client clusters is then treated as an isolated federated learning problem where either FedAvg or our proposed FedAP is employed. The full pipeline is visualized in \autoref{fig:pipeline}.

\section{Experiments and Results}
In this section, we present the experimental setup and the results of our experiments. The dataset, data preprocessing, and the employed data split for federated learning are discussed in \autoref{ex:ss1} and \autoref{ex:ss2}, respectively. We compare our proposed personalized clustered models to two baselines, a standard supervised model trained in the centralized setting and FedAvg \cite{mcmahan2017}. Furthermore, we present the results of combining clustering and personalization with existing approaches on the HAM10k skin lesion dataset \cite{tschandl2018}. 

\subsection{Experimental Setup}
We employ a pretrained MobileNetV2 \cite{sandler2018,sae2019} on ImageNet as a base classifier for all the baselines and our proposed model.
The hyperparameter tuning was performed using a validation set without any overlap with the test set for the centralized model.
For all the federated learning experiments, hyperparameter optimization comprised two steps: First, all clients were generated using a fixed random seed, and the hyperparameters were optimized against the client's test sets. Then, we generated all clients anew for the evaluation using a different random seed, and a final learning curve was acquired using the previously fixed hyperparameter. The following hyperparameters were optimized: learning rate, batch size, and a number of local training epochs. The final values were re-used in all of the experiments. The models were optimized with SGD optimizer, with learning rate $0.001$, inner epochs $e=1$, inner personalization epochs $7$, inner batch size $16$, initial and final meta-learning rate $\eta_0 = 1.0, \eta_k = 0.46$, total federated rounds $k = 220$, meta batch size $n = 5$, cluster initialization rounds $20$, Euclidean distance metric, ward linkage mechanism and maximum distance of $5$. 
\subsubsection{Dataset and preprocessing} \label{ex:ss1}
The HAM10k dataset \cite{tschandl2018} is a collection of $10,015$ dermatoscopic images of seven types of skin lesions.
The ground truth labels in HAM10k are based on histopathology in over 50\% of the cases and on follow-up examination, expert consensus, or in-vivo confocal microscopy. Our choice of HAM10k for evaluation of our method was due to this dataset's unbalanced and high non-i.i.d nature. 
To address the problem of unbalancedness in the dataset, we utilized random undersampling \cite{fernandez2018learning}; \ie at most 500 images from each lesion class were randomly sampled and used for training.

\subsubsection{Federated data split} \label{ex:ss2}
One of the common problems in clinical datasets that challenge machine learning methods is low statistical heterogeneity \cite{wynants2018random}. We modeled highly non-i.i.d data distributions between our clients to represent this problem. Images within each class were partitioned into 35 groups, and the clients were randomly assigned two partitions from different classes until no partition pairs were left. This resulted in 34 clients, with 70 images assigned to each in total from two classes (see the supplementary material for the distribution heatmap). The 70 images were randomly split into training and test sets at an 80:20 ratio within each client. 

\subsection{Results and Discussions}
In this section, we present the results of our experiments and the comparison of our proposed model to previous work. In order to take the number of personalization rounds in FedAP and initialization rounds in HC  into account for our total training rounds, the reported accuracy values in \autoref{ex:tbl1} are based on the total number of training steps for all models.

\begin{minipage}[b]{0.46\textwidth}
    \centering
    \includegraphics[width=\linewidth]{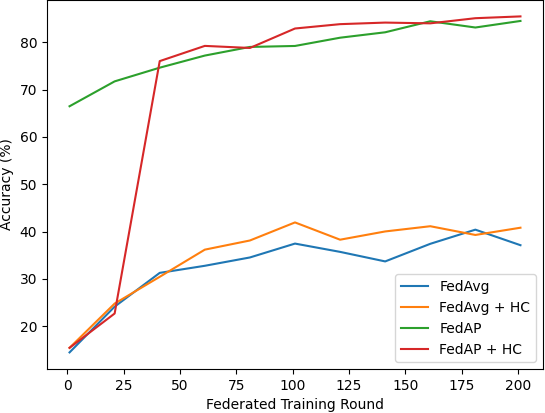}
    \captionof{figure}{\textbf{Average Test Accuracy in Different Rounds}. Classification accuracy averaged over all clients. This figure shows the increase in the convergence rate using both FedAP and HC.} 
    \label{fig:acc}
  \end{minipage}
  \hspace{0.05cm}
  \begin{minipage}[b]{0.46\textwidth}
    \centering
    \begin{tabular}{lc}\hline
\hline
Method & Accuracy (\%)\\
\hline\hline
Centralized & 76.8 \\ 
\hline
FedAvg \cite{mcmahan2017} & 41.1 $\pm$ 34.31 \\ 
FedAvg + HC \cite{briggs2020federated} & 44.1 $\pm$ 20.86\\ 
FedAP (Ours) & 84.1 $\pm$ 14.53 \\ 
FedAP + HC (Ours) & \textbf{86.9} $\pm$ \textbf{12.81}\\ 
\hline
      \end{tabular}
      \captionsetup{skip=35pt}

      \captionof{table}{\textbf{Results}. The comparison of our methods to the baselines on the non-i.i.d HAM10k dataset. The reported values for federated models are based on the mean and std of all the clients.}
    \label{ex:tbl1}
    \end{minipage}

A standard \textbf{centralized} training setting was intended as a standard against which the more constrained federated learning experiments could be measured. For this experiment, a global training and test sets were created by pooling all clients' respective training and test data. Due to the availability of all clients' data to a single model, this model achieves relatively higher performance compared to FedAvg.

\textbf{FedAvg} achieves the worst performance across all methods. The learning curve shows a high degree of oscillation due to the fact that the single global model was not able to learn all necessary features from the different learning tasks imposed by different data distributions of the clients. Thus, this demonstrates the lack of robustness of \textit{FedAvg} in a highly non-i.i.d. data setting, as indicated by preliminary evidence \cite{zhao2018federated,li_convergence_2020,sattler_robust_2019}.

For the \textbf{FedAvg + HC} experiment, several FedAvg rounds (given by the hyperparameter "cluster initialization rounds") were performed before clustering. Following this, clients were clustered according to their local updates using hierarchical clustering (as in Briggs \etal \cite{briggs2020federated}). Inside the individual clusters, conventional \textit{FedAvg} training was performed. The global accuracy was obtained by simply averaging the test set accuracy of all clients. Hierarchical clustering increased the overall convergence speed of \textit{FedAvg}. A sudden jump in the accuracy, as well as final model performance improvement, can be seen after clustering in \autoref{fig:acc}. We assume that an approximation of a homogeneous learning setting is recovered inside the clusters, and indeed the model updates can be used to represent the data distribution. However, this approach still falls far behind the performance of \textit{FedAP}, substantiating the crucial importance of model personalization in a highly non-i.i.d. data environment. 
similar to \textit{FedAvg}, \textbf{FedAP} randomly samples a batch of $n$ clients at each round and uses SGD for local training. SGD is also used for global training so that the global update rule became $\theta \gets \theta + \eta \cdot \Delta \theta$. Moreover, the adaptive weight decreases linearly with the number of federated rounds, from specified initial to final values. The evaluation of \textit{FedAP} was performed in the same way as of FedAvg with the difference that, after the distribution of the global model, it was personalized to the clients by performing a number of gradient optimization steps on the local training sets. After this personalization, the model was evaluated on the test set, and again a global accuracy was calculated by averaging all local test set accuracies. The global model for \textit{FedAP} was able to learn continuously, while the personalization step allowed the adaption to the different client distributions. Especially the extreme non-i.i.d. setting with simple underlying tasks allows the personalization to perform very well.

The \textbf{FedAP + HC} experiment was performed in the same way as the previous one, only that the adaptive personalization is performed after hierarchical clustering. As in the \textit{FedAvg + HC} experiment, clusters were formed according to clients' model updates after initial $20$ rounds of FedAvg. \textit{FedAP + HC} shows the overall best performance compared to all other approaches. We observe the same learning curve behavior as in \textit{FedAvg + HC}. It improves the convergence speed of the algorithm, shown again by the immediate jump in balanced accuracy after clustering depicted in \autoref{fig:acc}.
Despite achieving the best performance among the different mentioned variations of the federated setting, we observed that the FedAP + HC model suffers from overfitting in very long runs; \ie if we train both FedAP and FedAP + HC for around 500 more rounds, the performance stays the same or gains minimal improvement in FedAP, but the FedAP + HC starts deteriorating. We assume that this is an indication of the global model's higher adaptability to distinct tasks than the cluster models. The core idea of meta-learning roots in learning from many distinctive tasks. Indeed, clustering clients (representing the meta-learning task in a federated learning setting) reduces the diversity of tasks in each cluster to learn from. Therefore, this might be the reason behind FedAP + HC model's sensitivity to overfitting in more extended training.
\section{Conclusion}
In this work, we presented and analyzed an adaptive personalization approach along with hierarchical clustering of the clients to tackle the non-i.i.d. problem in federated learning. Our adaptive model parameter weighting with hierarchical clustering enables the better adaptation of the local models of clients to their distinctive data distribution while still taking advantage of the global aggregation of the different client model updates. The clients are clustered based on their similarity to local model updates and thus can approximate an i.i.d. setting in the respective clusters. Our experiments on the HAM10k dataset with the MobileNetV2 network show drastic improvements in classification accuracy over the standard supervised and over the FedAvg baseline. The lower standard deviation of our proposed method compared to previous work demonstrate that adaptive personalization of client models in the federated setting, inspired by meta-learning, yields higher generalizability of all clients models. In addition, hierarchical clustering increases the convergence speed and allows for better global models. Our experiments show that the models trained with FedAP and HC have the lowest standard deviation and highest average accuracy, demonstrating the proposed methods' effectiveness in reaching a reasonable and high accuracy performance in all clients. Despite the high performance gain of our proposed method, if the model is trained for too many rounds, the performance decreases, which shows its sensitivity to overfitting. Therefore, we plan to investigate this issue in future work.

\section*{Acknowledgement}
This work was supported in part by the Munich Center for Machine Learning (MCML) with funding from the Bundesministerium f\"ur Bildung und Forschung (BMBF) under the project 01IS18036B.

\bibliographystyle{splncs04}
\bibliography{main}

\begin{thebibliography}{10}
\providecommand{\url}[1]{\texttt{#1}}
\providecommand{\urlprefix}{URL }
\providecommand{\doi}[1]{https://doi.org/#1}

\bibitem{aliakbari2015simulation}
Aliakbari, A., Yeganeh, Y.M., Safari, S.: Simulation of dac-based truncated
  sine excitation pulse generator. In: 2015 2nd International Conference on
  Knowledge-Based Engineering and Innovation (KBEI). pp. 689--693. IEEE (2015)

\bibitem{arivazhagan_federated_2019}
Arivazhagan, M.G., Aggarwal, V., Singh, A.K., Choudhary, S.: Federated
  {Learning} with {Personalization} {Layers}. arXiv:1912.00818 [cs, stat]  (Dec
  2019), \url{http://arxiv.org/abs/1912.00818}, arXiv: 1912.00818

\bibitem{briggs2020federated}
Briggs, C., Fan, Z., Andras, P.: Federated learning with hierarchical
  clustering of local updates to improve training on non-iid data (May 2020)

\bibitem{chen2019federated}
Chen, F., Luo, M., Dong, Z., Li, Z., He, X.: Federated meta-learning with fast
  convergence and efficient communication (2019)

\bibitem{fallah2020personalized}
Fallah, A., Mokhtari, A., Ozdaglar, A.: Personalized federated learning: A
  meta-learning approach (2020)

\bibitem{farshad2022upsilon}
Farshad, A., Yeganeh, Y., Gehlbach, P., Navab, N.: Y-net: A spatiospectral
  dual-encoder networkfor medical image segmentation. arXiv preprint
  arXiv:2204.07613  (2022)

\bibitem{fernandez2018learning}
Fern{\'a}ndez, A., Garc{\'\i}a, S., Galar, M., Prati, R.C., Krawczyk, B.,
  Herrera, F.: Learning from imbalanced data sets, vol.~11. Springer (2018)

\bibitem{finn2017}
Finn, C., Abbeel, P., Levine, S.: Model-agnostic meta-learning for fast
  adaptation of deep networks. In: Precup, D., Teh, Y.W. (eds.) Proceedings of
  the 34th International Conference on Machine Learning. Proceedings of Machine
  Learning Research, vol.~70, pp. 1126--1135. PMLR, International Convention
  Centre, Sydney, Australia (06--11 Aug 2017),
  \url{http://proceedings.mlr.press/v70/finn17a.html}

\bibitem{jiang2019improving}
Jiang, Y., Konecny, J., Rush, K., Kannan, S.: Improving federated learning
  personalization via model agnostic meta learning (2019)

\bibitem{khodak2019adaptive}
Khodak, M., Balcan, M.F., Talwalkar, A.: Adaptive gradient-based meta-learning
  methods (2019)

\bibitem{li_fedmd_2019}
Li, D., Wang, J.: {FedMD}: {Heterogenous} {Federated} {Learning} via {Model}
  {Distillation}. arXiv:1910.03581 [cs, stat]  (Oct 2019),
  \url{http://arxiv.org/abs/1910.03581}, arXiv: 1910.03581

\bibitem{li2020federated}
Li, T., Sahu, A.K., Talwalkar, A., Smith, V.: Federated learning: Challenges,
  methods, and future directions. IEEE Signal Processing Magazine
  \textbf{37}(3),  50--60 (2020)

\bibitem{li_convergence_2020}
Li, X., Huang, K., Yang, W., Wang, S., Zhang, Z.: On the {Convergence} of
  {FedAvg} on {Non}-{IID} {Data}. arXiv:1907.02189 [cs, math, stat]  (Jun
  2020), \url{http://arxiv.org/abs/1907.02189}, arXiv: 1907.02189

\bibitem{madabhushi2016image}
Madabhushi, A., Lee, G.: Image analysis and machine learning in digital
  pathology: Challenges and opportunities. Medical Image Analysis  \textbf{33},
   170--175 (2016). \doi{https://doi.org/10.1016/j.media.2016.06.037},
  \url{https://www.sciencedirect.com/science/article/pii/S1361841516301141},
  20th anniversary of the Medical Image Analysis journal (MedIA)

\bibitem{mcmahan2017}
McMahan, B., Moore, E., Ramage, D., Hampson, S., y~Arcas, B.A.:
  Communication-efficient learning of deep networks from decentralized data.
  In: Singh, A., Zhu, J. (eds.) Proceedings of the 20th International
  Conference on Artificial Intelligence and Statistics. Proceedings of Machine
  Learning Research, vol.~54, pp. 1273--1282. PMLR, Fort Lauderdale, FL, USA
  (20--22 Apr 2017), \url{http://proceedings.mlr.press/v54/mcmahan17a.html}

\bibitem{nichol2018}
Nichol, A., Achiam, J., Schulman, J.: On first-order meta-learning algorithms
  (2018)

\bibitem{reddi_adaptive_2020}
Reddi, S., Charles, Z., Zaheer, M., Garrett, Z., Rush, K., Konečný, J.,
  Kumar, S., McMahan, H.B.: Adaptive {Federated} {Optimization}.
  arXiv:2003.00295 [cs, math, stat]  (Dec 2020),
  \url{http://arxiv.org/abs/2003.00295}, arXiv: 2003.00295

\bibitem{rieke2021future}
Rieke, N., Hancox, J., Li, W., Milletari, F., Roth, H., Albarqouni, S., Bakas,
  S., Galtier, M.N., Landman, B., Maier-Hein, K., Ourselin, S., Sheller, M.,
  Summers, R.M., Trask, A., Xu, D., Baust, M., Cardoso, M.J.: The future of
  digital health with federated learning (2021)

\bibitem{sae2019}
Sae-Lim, W., Wettayaprasit, W., Aiyarak, P.: Convolutional neural networks
  using mobilenet for skin lesion classification. In: 2019 16th international
  joint conference on computer science and software engineering (JCSSE). pp.
  242--247. IEEE (2019)

\bibitem{sandler2018}
Sandler, M., Howard, A., Zhu, M., Zhmoginov, A., Chen, L.C.: Mobilenetv2:
  Inverted residuals and linear bottlenecks. In: Proceedings of the IEEE
  conference on computer vision and pattern recognition. pp. 4510--4520 (2018)

\bibitem{sattler_robust_2019}
Sattler, F., Wiedemann, S., Müller, K.R., Samek, W.: Robust and
  {Communication}-{Efficient} {Federated} {Learning} from {Non}-{IID} {Data}.
  arXiv:1903.02891 [cs, stat]  (Mar 2019),
  \url{http://arxiv.org/abs/1903.02891}, arXiv: 1903.02891

\bibitem{sheller2020federated}
Sheller, M.J., Edwards, B., Reina, G.A., Martin, J., Pati, S., Kotrotsou, A.,
  Milchenko, M., Xu, W., Marcus, D., Colen, R.R., et~al.: Federated learning in
  medicine: facilitating multi-institutional collaborations without sharing
  patient data. Scientific reports  \textbf{10}(1),  1--12 (2020)

\bibitem{smith_federated_2018}
Smith, V., Chiang, C.K., Sanjabi, M., Talwalkar, A.: Federated {Multi}-{Task}
  {Learning}. arXiv:1705.10467 [cs, stat]  (Feb 2018),
  \url{http://arxiv.org/abs/1705.10467}, arXiv: 1705.10467

\bibitem{tschandl2018}
Tschandl, P., Rosendahl, C., Kittler, H.: The ham10000 dataset, a large
  collection of multi-source dermatoscopic images of common pigmented skin
  lesions. Scientific data  \textbf{5}(1), ~1--9 (2018)

\bibitem{vanschoren2018meta}
Vanschoren, J.: Meta-learning: A survey. arXiv preprint arXiv:1810.03548
  (2018)

\bibitem{wang_federated_2019}
Wang, K., Mathews, R., Kiddon, C., Eichner, H., Beaufays, F., Ramage, D.:
  Federated {Evaluation} of {On}-device {Personalization}. arXiv:1910.10252
  [cs, stat]  (Oct 2019), \url{http://arxiv.org/abs/1910.10252}, arXiv:
  1910.10252

\bibitem{wynants2018random}
Wynants, L., Riley, R., Timmerman, D., Van~Calster, B.: Random-effects
  meta-analysis of the clinical utility of tests and prediction models.
  Statistics in medicine  \textbf{37}(12),  2034--2052 (2018)

\bibitem{yeganeh_inverse_2020}
Yeganeh, Y., Farshad, A., Navab, N., Albarqouni, S.: Inverse {Distance}
  {Aggregation} for {Federated} {Learning} with {Non}-{IID} {Data}.
  arXiv:2008.07665 [cs, stat]  (Aug 2020),
  \url{http://arxiv.org/abs/2008.07665}, arXiv: 2008.07665

\bibitem{yue_deep_2020}
Yue, L., Tian, D., Chen, W., Han, X., Yin, M.: Deep learning for heterogeneous
  medical data analysis. World Wide Web  \textbf{23}(5),  2715--2737 (Sep
  2020). \doi{10.1007/s11280-019-00764-z},
  \url{https://doi.org/10.1007/s11280-019-00764-z}

\bibitem{zhao2018federated}
Zhao, Y., Li, M., Lai, L., Suda, N., Civin, D., Chandra, V.: Federated learning
  with non-iid data. CoRR  \textbf{abs/1806.00582} (2018),
  \url{http://arxiv.org/abs/1806.00582}

\end{thebibliography}

\end{document}